\documentclass{article}
\usepackage{graphicx}
\usepackage{amsfonts}
\usepackage{dsfont}
\usepackage{amssymb}
\usepackage{amsmath}
\usepackage{amsthm}
\usepackage{epsf}
\usepackage{dsfont}
\usepackage{color}
\usepackage{multirow}
\usepackage{epsfig}
\def\Tcal{\mathcal{T}}

\def\Xcal{\mathcal{X}}

\def\Mscal{M_+^s(\Xcal)}

\def\Pcal{\mathcal{P}}

\def\Mtcal{M_\Tcal(\Xcal)}

\DeclareMathOperator{\defi}{def}

\DeclareMathOperator{\defeq}{\overset{\defi}{=}}

\DeclareMathOperator{\card}{card}

\DeclareMathOperator{\ins}{\in\,}

\def\OMIT#1{}

\theoremstyle{break} 
\theoremstyle{break} \newtheorem{definition}{Definition}
\theoremstyle{break}
\newtheorem{theorem}{Theorem} \theoremstyle{marginbreak}

\begin{document}

\title{Multiresolution Kernels}
\author{
Marco Cuturi\\
Ecole des Mines de Paris \\
Fontainebleau, France;\\
Institute of Statistical Mathematics\\
Tokyo, Japan.\\
\texttt{marco.cuturi@ensmp.fr} \and
Kenji Fukumizu\\
Institute of Statistical Mathematics\\
Tokyo, Japan.\\
\texttt{fukumizu@ism.ac.jp} }
\date{June 5th, 2005}
\maketitle

\begin{abstract}
We present in this work a new methodology to design kernels on
data which is structured with smaller components, such as text,
images or sequences. This methodology is a template procedure
which can be applied on most kernels on measures and takes
advantage of a more detailed ``bag of components'' representation
of the objects. To obtain such a detailed description, we consider
possible decompositions of the original bag into a collection of
nested bags, following a prior knowledge on the objects'
structure. We then consider these smaller bags to compare two
objects both in a detailed perspective, stressing local matches
between the smaller bags, and in a global or coarse perspective,
by considering the entire bag. This multiresolution approach is
likely to be best suited for tasks where the coarse approach is
not precise enough, and where a more subtle mixture of both local
and global similarities is necessary to compare objects. The
approach presented here would not be computationally tractable
without a factorization trick that we introduce before presenting
promising results on an image retrieval task.
\end{abstract}

\section{Introduction}\label{sec:intro}
There is strong evidence that kernel methods~\cite
{schoelkopf02learning} can deliver state-of-the-art performance on
most classification tasks when the input data lies in a vector
space. Arguably, two factors contribute to this success. First,
the good ability of kernel algorithms, such as the SVM, to
generalize and provide a sparse formulation for the underlying
learning problem; Second, the capacity of nonlinear kernels, such
as the polynomial and RBF kernels, to quantify meaningful
similarities between vectors, notably non-linear correlations
between their components. Using kernel machines with non-vectorial
data (e.g., in bioinformatics, pattern recognition or signal
processing tasks) requires more arbitrary choices, both to
represent the objects and to chose suitable kernels on those
representations. The challenge of using kernel methods on
real-world data has thus recently fostered many proposals for
kernels on complex objects, notably for strings, trees, images or
graphs to cite a few.

\par A strategy often quoted as the generative approach to this problem
takes advantage of a generative model, that is an adequate
statistical model for the objects, to derive feature
representations for the objects. In practice this often yields
kernels to be used on the histograms of smaller components sampled
in the objects, where the kernels take into account the geometry
of the underlying model in their similarity measures~\cite
{jebara04probability,Lafferty2005,cuturi05context,hein05hilbertian,cuturi05semigroup}.
The previous approaches coupled with SVM's combine both the
advantages of using discriminative methods with generative ones,
and produced convincing results on many tasks.

\par One of the drawbacks of such representations is however that they implicitly assume
that each component has been generated independently and in a
stationary way, where the empirical histogram of components is
seen as a sample from an underlying stationary measure. While this
viewpoint may translate into adequate properties for some learning
tasks (such as translation or rotation invariance when using
histograms of colors to manipulate images \cite{chapelle99svms}),
it might prove too restrictive and hence inadequate for other
types of problems. Namely, tasks which involve a more subtle mix
of detecting \emph{both} conditional (with respect to the location
of the components for instance) and global similarities between
the objects. Such problems are likely to arise for instance in
speech, language, time series or image processing. In the first
three tasks, this consideration is notably treated by most
state-of-the-art methods through dynamic programming algorithms
capable of detecting and penalizing accordingly local matches
between the objects. Using dynamic programming to produce a kernel
yielded fruitful results in different
applications~\cite{VerSaiAku04,shimodaira02dynamic}, with the
limitation that the kernels obtained in practice are not always
positive definite, as reviewed in~\cite{VerSaiAku04}. Other
kernels proposed for sequences~\cite{RaeSon04} directly
incorporate a localization information into each component,
augmenting considerably the size of the component space, and then
introduce some smoothing (such as mismatches) to avoid
representations that would be too sparse.

\begin{figure}[htbp]
\begin{center}
\scalebox{.4}{\input{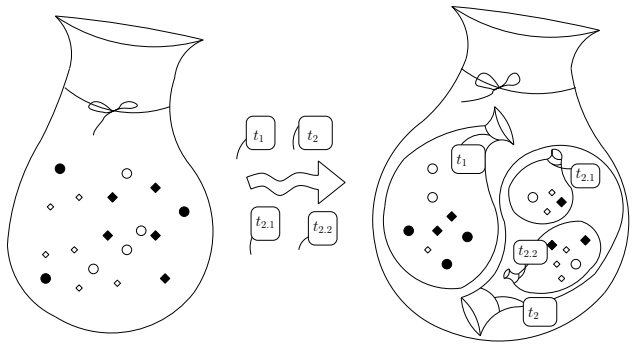}}
\caption{From the bag of components representation to a set of nested bags, using a set of conditioning events.} \label{fig:bags}
\end{center}
\end{figure}

\par We propose in this work a different approach grounded on the generative approach previously quoted, managing however to combine both conditional and global similarities when comparing two objects. The motivation behind this approach is both intuitive
and computational: intuitively, the global histogram of
components, that is the simple bag of components representation of
Figure~\ref{fig:bags}, may seem inadequate if the components'
appearance seem to be clearly conditioned by some external events.
This phenomenon can be taken into account by considering
collections (indexed on the same set of events, to be defined) of
nested bags or histograms to describe the object. Kernels that
would only rely on these detailed resolutions might however miss
the bigger picture that is provided by the global histogram. We
propose a trade-off between both viewpoints through a combination
that aims at giving a balanced account of both fine and coarse
perspectives, hence the name of multiresolution kernels, which we
introduce formally in Section~\ref{sec:multi}. On the
computational side, we show how such a theoretical framework can
translate into an efficient factorization detailed in
Section~\ref{sec:factor}. We then provide experimental results in
Section~\ref{sec:exp} on an image retrieval task which shows that
the methodology improves the performance of kernel based
state-of-the art techniques in this field.

\section{Multiresolution Kernels}\label{sec:multi}
In most applications, complex objects can be represented as histograms of components,
such as texts as bags of words or images and sequences as histograms of colors and letters. Through this
representation, objects are cast as probability laws or measures
on the space $\Xcal$ of components, typically multinomials if $\Xcal$ is finite
~\cite{Lafferty2005,hein05hilbertian,chapelle99svms,joachims:2002a},
and compared as such through kernels on measures. An obvious drawback of this representation is that all
contextual information on how the components have been sampled
is lost, notably any general sense of position in the objects, but
also more complex conditional information that may be induced from
neighboring components, such as transitions or long range
interactions.

\par In the case of images for instance, one may be tempted to consider not only the overall histogram of colors, but also
more specialized histograms which may be relevant for the task. If some
local color-overlapping in the images is an interesting or decisive
feature of the learning problem, these specialized histograms may be generated arbitrarily following a grid,
dividing for instance the image into 4 equal parts, and computing
histograms for each corner before comparing them pairwise between two
images (see Figure~\ref{fig:multiresgrid} for an illustration).
If sequences are at stake, these may also be sliced into predefined regions to yield local histograms of letters. If the strings are on the contrary assumed to follow some Markovian behaviour (namely that the appearance of letters in the string is independent of their exact location but only depends on the few letters that precede them), an interesting index would translate into a set of contexts, typically a complete suffix dictionary as detailed in~\cite{cuturi05context}. While the two
previous examples may seem opposed in the way the histograms are
generated, both methodologies stress a particular class of events (location or transitions) that give an additional knowledge on how the components were sampled in the objects.
Since both these two approaches, and possibly other ones, can be applied within the framework of this paper using a unified formalism, we
present our methodology using a general notation for the index
of events. Namely, we note $\Tcal$ for an arbitrary set of conditioning events, assuming these events can be
directly observed on the object itself, by contrast with the
latent variables approach of ~\cite{tsuda02marginalized}. Considering still, following the generative approach, that an object can be mapped onto a probability measure $\mu$ on $\Xcal$, we have that the
realization of an event $t\ins\Tcal$ can be interpreted under the light of a joint probability
$\mu(x,t)$, with $x\ins\Xcal$, factorized through Bayes'
law as $\mu(x|t)\mu(t)$ to yield the following decomposition of
$\mu$ as
$$
\mu=\sum_{t\ins\Tcal}\mu_t,
$$
where each $\mu_t\defeq\mu(\cdot|t)\mu(t)$ is an element of the set of sub-probability measures $\Mscal$, that is the set of positive measures $\rho$ on
$\Xcal$ such that their total mass $\rho(\Xcal)$ denoted as
$|\rho|$ is \emph{less than} or equal to $1$. To take into account
the information brought by the events in $\Tcal$, objects can
hence be represented as families of measures of $\Mscal$ indexed by
$\Tcal$, namely elements $\mu$ contained in $\Mtcal\overset{\defi}{=}\Mscal^\Tcal.$

\subsection{Local Similarities Between Measures Conditioned by Sets of Events}
To compare two objects under the light of their respective
decompositions as sub-probability measures $\mu_t$ and $\mu_t'$, we make
use of an arbitrary positive definite kernel $k$ on $\Mscal$ to which we
will refer to as the base kernel throughout the paper. For interpretation
purposes only, we may assume in the following sections that $k$ can be written as $e^{-d^2}$
 where $d$ is an Euclidian distance in $\Mscal$. Note also that the kernel is defined not only on probability
measures, but also on sub-probabilities. For two elements $\mu,\mu'$
of $\Mtcal$ and a given element $t\ins\Tcal$, the kernel
$$k_t(\mu,\mu')\overset{\defi}{=}k(\mu_t,\mu'_t)$$ measures the
similarity of $\mu$ and $\mu'$ by quantifying how similarly their components
were generated conditionally to event $t$. For two different events
$s$ and $t$ of $\Tcal$, $k_s$ and $k_t$ can be associated through
polynomial combinations with positive factors to result in new
kernels, notably their sum $k_s+k_t$ or their product $k_s k_t$. This is particularly adequate if
some complementarity is assumed between $s$ and $t$, so that their
combination can provide new insights for a given learning task. If
on the contrary the events are assumed to be similar, then they
can be regarded as a unique event $\{s\}\cup\{t\}$ and result in
the kernel
$$k_{\{s\}\cup\{t\}}(\mu,\mu')\defeq k(\mu_s+\mu_t,\mu'_s+\mu'_t),$$
which will measure the similarity of $m$ and $m'$ when \emph{either} $s$
or $t$ occurs. The previous formula can be extended to model
kernels indexed on a set $T\subset\Tcal$ of similar events, through
$$k_{T}(m,m')\defeq k\left(\mu_T,\mu'_T\right), \;\text{where }\;\mu_T\defeq\sum_{t\ins T}{\mu_t}\;\text{ and }\;\mu'_T\defeq\sum_{t\ins T}{\mu'_t}.$$
Note that this equivalent to defining a distance between elements $\mu$
and $\mu'$ conditionned by $T$ as
$d^2_T(\mu,\mu')\defeq d^2(\mu_T,\mu'_T)$.

\subsection{Resolution Specific Kernels}
Let $P$ be a finite partition of $\Tcal$, that is a finite family
$P=(T_1,...,T_n)$ of sets of $\Tcal$, such that $T_i\cap T_j=\varnothing$ if $1\leq i<j\leq n$ and $\bigcup_{i=1}^n T_i=\Tcal$. We write
$\Pcal(\Tcal)$ for the set of all partitions of $\Tcal$. Consider
now the kernel defined by a partition $P$ as
\begin{equation}\label{eq:partitionspecific}
k_P(\mu,\mu')\defeq\prod_{i=1}^n k_{T_i}(\mu,\mu').
\end{equation} The kernel $k_P$ quantifies the
similarity between two objects by detecting their joint similarity under all possible events of
$\Tcal$, given an a priori similarity assumed on the events which is
expressed as a partition of $\Tcal$. Note that there is some arbitrary
in this definition since, following the convolution
kernels~\cite{haussler99convolution} approach for instance, a simple multiplication of base
kernels $k_{T_i}$ to define $k_P$ is used, rather than any other polynomial
combination. More precisely, the multiplicative
structure of Equation~\eqref {eq:partitionspecific} quantifies how two objects are similar given a
partition $P$ in a way that imposes for the objects to be similar
according to all subsets $T_i$. If $k$ can be expressed as a function of a distance $d$, $k_P$ can be expressed as the exponential of
$$d^2_P(\mu,\mu')\defeq \sum_{i=1}^n d^2_{T_i}(\mu,\mu'),$$ a quantity which penalizes
local differences between the decompositions of $\mu$ and $\mu'$ over $\Tcal$, as opposed
to the coarsest approach where $P=\{\Tcal\}$ and only $d^2(\mu,\mu')$ is
considered.
\begin{figure}[htbp]
\begin{center}
\includegraphics[width=12cm]{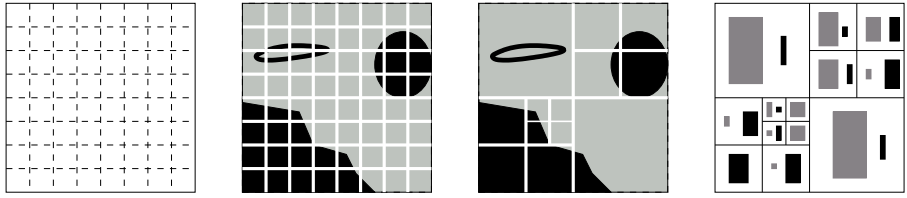} \caption{A useful
set of events $\Tcal$ for images which would focus on pixel
localization can be represented by a grid, such as the $8\times 8$
one represented above. In this case $P_3$ corresponds to the $4^3$ windows presented in the left image, $P_2$ to the $16$ larger square obtained when grouping $4$ small windows, $P_1$ to the image divided into $4$ equal parts and $P_0$ is simply the whole image. Any partition of the image obtained from sets in $P_0^3$, such as the one represented above, can in turn be used to represent an image
as a family of sub-probability measures, which reduces in the case
of two-color images to binary histograms as illustrated in the
right-most image.}\label{fig:multiresgrid}
\end{center}
\end{figure}
\par As illustrated in Figure~\ref{fig:multiresgrid} in the case of
images expressed as histograms indexed over locations, a partition of $\Tcal$
reflects a given belief on how events should be associated to belong to the same set or dissociated to
highlight interesting dissimilarities. Hence, all partitions
contained in the set $\Pcal(\Tcal)$ of all possible
partitions\footnote{which is quite a big space, since if $\Tcal$
is a finite set of cardinal $r$, the cardinal of the set of
partitions is known as the Bell Number of order $r$ with
$B_r=\frac{1}{e}\sum_{u=1}^{\infty}\frac{u^r}{u!}\underset{r\rightarrow\infty}{\sim}
e^{r\ln r}$.} are not likely to be equally meaningful given that some events may look more similar than others. If the index is based on location, one would naturally favor mergers between neighboring indexes. For contexts, a useful topology might also be derived by grouping contexts with similar suffixes.
\par Such meaningful partitions can be obtained in a general case if we assume the existence of a prior hierarchical information on the elements of $\Tcal$, translated into a series
$$P_0=\{\Tcal\},..,P_D=\{\{t\},t\ins\Tcal\}$$ of partitions of
$\Tcal$, namely a hierarchy on $\Tcal$. To provide a hierarchical content, the family $(P_d)_{d=1}^{D}$ is such that any subset
present in a partition $P_d$ is included in a (unique by
definition of a partition) subset included in the coarser
partition $P_{d-1}$, and further assume this inclusion to be
strict. This is equivalent to stating that each set $T$ of a
partition $P_d$ is divided in $P_{d+1}$ through a partition of $T$
 which is not $T$ itself. We note this partition $s(T)$ and name its
 elements the siblings of $T$. Consider now the subset $\Pcal_D\subset\Pcal(\Tcal)$ of all partitions of
$\Tcal$ obtained by using only sets in $$P_0^D\overset{\defi}{=}\bigcup_{d=1}^{D} P_d,$$ namely $
\Pcal_D\overset{\defi}{=}\{P\ins\Pcal(\Tcal) \text{ s.t. } \forall \,T \ins P, T\ins P_0^D\}.$. The set $\Pcal_D$ contains both the coarsest and the finest resolutions, respectively $P_0$ and $P_D$, but also all variable resolutions
for sets enumerated in $P_0^D$, as can be seen for instance in the third
image of Figure~\ref{fig:multiresgrid}.

\subsection{Averaging Resolution Specific Kernels}
Each partition $P$ contained in $\Pcal_D$ provides a resolution to
compare two objects, and generates consequently a very large
family of kernels $k_P$ when $P$ spans $\Pcal_D$. Some partitions
are probably better suited for certain tasks than others, which
may call for an efficient estimation of an optimal partition given
a task. We take in this section a different direction by
considering an averaging of such kernels based on a Bayesian prior
on the set of partitions. In practice, this averaging favours
objects which share similarities under a large collection of
resolutions.
\begin{definition}\label{def:multiresolution}
Let $\Tcal$ be an index set endowed with a hierarchy $(P_d)_{d=0}^{D}$, $\pi$ be a prior measure on the corresponding set of partitions $\Pcal_D$ and $k$ a base kernel on $\Mscal\times\Mscal$. The multiresolution kernel $k_\pi$ on $\Mtcal\times\Mtcal$ is defined as
\begin{equation}\label{eq:multiresolution}
k_\pi(\mu,\mu')=\sum_{P\ins\Pcal_D}\pi(P)\, k_P(\mu,\mu').
\end{equation}
\end{definition}
Note that in Equation~\eqref{eq:multiresolution}, each resolution specific
kernel contributes to the final kernel value and may be
regarded as a weighted feature extractor.
\section{Kernel Computation}\label{sec:factor}
This section aims at characterizing hierarchies $(P_d)_{d=0}^{D}$
and priors $\pi$ for which the computation of $k_\pi$ is both
tractable and meaningful. We first propose a type of hierarchy
generated by trees, which is then coupled with a branching process
prior to fully specify $\pi$. These settings yield a computational
time for expressing $k_\pi$ which is loosely upperbounded by
$D\times\card{\Tcal}\times c(k)$ where $c(k)$ is the time required
to compute the base kernel.
\subsection{Partitions Generated by Branching Processes}\label{sec:priors}
All partitions $P$ of $\Pcal_D$ can be generated iteratively through the following rule,
starting from the initial root partition $P:=P_0=\{\Tcal\}$. For each set $T$ of $P$:
\begin{enumerate}\item either leave the set as it is in $P$,
\item either replace it by its
siblings enumerated in $s(T)$, and
reapply this rule to each sibling unless they belong to the finest partition $P_D$.
\end{enumerate}
By giving a probabilistic content to the previous rule through a
binomial parameter (i.e. for each treated set assign probability
$1-\varepsilon$ of applying rule~1 and probability $\varepsilon$
of applying rule~2) a candidate prior for $\Pcal_D$ can be derived, depending on the overall coarseness of the considered partition.
For all elements $T$ of $P_D$ this binomial parameter is equal to
$0$, whereas it can be individually defined for any element $T$ of
the $D-1$ coarsest partitions as $\varepsilon_T\ins[0,1]$, yielding for a partition $P\ins\Pcal_D$ the weight
$$
\pi(P)=\prod_{T\ins P}(1-\varepsilon_T)\prod_{T\ins \overset{\circ}{P}}(\varepsilon_T),
$$
where the set $\overset{\circ}{P}=\{T\ins P_0^D \text{ s.t. }
\exists V\ins P, V \subsetneq T \}$ gathers all coarser sets
belonging to coarser resolutions than $P$, and can be regarded as
all ancestors in $P_0^D$ of sets enumerated in $P$.
\subsection{Factorization}
\begin{figure}[htbp]
\begin{center}
\scalebox{.5}{\input{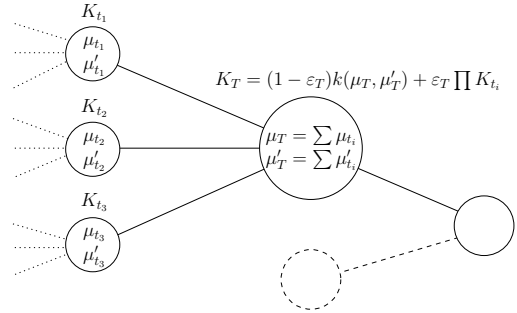}}
\caption{The update rule for the computation of $k\pi$ takes into account the branching process prior by updating each node corresponding to a set $T$ of any intermediary partitions with the values obtained for higher resolutions in $s(T)$.} \label{fig:fit}
\end{center}
\end{figure}

The prior proposed in Section~\ref{sec:priors} can be used to
factorize the formula in~\eqref{eq:multiresolution}, which is
summarized in this theorem, using notations used in
Definition~\ref{def:multiresolution}
\begin{theorem}
For two elements $m,m'$ of $\Mtcal$, define for $T$ spanning recursively
$P_D,P_{D-1},...,P_0\,$ the quantity $$K_T=(1-\varepsilon_T)k_T(\mu,\mu')+\varepsilon_T
\prod_{U\ins\, s(T)}K_U.$$
Then $k_\pi(\mu,\mu')=K_{\Tcal}$.
\end{theorem}
\begin{proof}
The proof follows from the prior structure used for the tree
generation, and can be found in either~\cite{stflour} or~\cite
{cuturi05context}. Figure~\ref{fig:fit} underlines the importance of
incorporating to each node $K_T$ a weighted product of the kernels $K_U$
computed by its siblings.
\end{proof}

If the hierarchy of $\Tcal$ is such that the cardinality of $s(T)$ is fixed
to a constant $\alpha$ for any set $T$, typically $\alpha=4$ for images
as seen in Figure~\ref{fig:multiresgrid}, then the computation of
$k_\pi$ is upperbounded by $(\alpha^{D+1}-1)c(k)$. This computational complexity may even
become lower in cases where the histograms become sparse at fine
resolutions, yielding complexities in linear time with respect to the
size of the compared objects, quantified by the length of the sequences in~\cite
{cuturi05context} for instance.
\section{Experiments}\label{sec:exp}
We present in this section experiments inspired by the image
retrieval task first considered in~\cite{chapelle99svms} and also
used in~\cite{hein05hilbertian}, although the images used here are
not exactly the same. The dataset was also extracted from the
Corel Stock database and includes 12 families of labelled images,
each class containing 100 color images, each image being coded as
$256\times384$ pixels with colors coded in 24 bits (16M colors).
The families depict \emph{bears, African specialty animals,
monkeys, cougars, fireworks, mountains, office interiors, bonsais,
sunsets, clouds, apes} and \emph{rocks and gems}. The database is
randomly split into balanced sets of 900 training images and 300
test images. The task consists in classifying the test images with
the rule learned by training 12 one-vs-all SVM's on the learning
fold. The object are then classified according to the SVM
performing the highest score, namely with a ``winner-takes-all''
strategy. The results presented in this section are averaged over
4 different random splits. We used the CImg package to generate
histograms and the Spider toolbox for the SVM
experiments\footnote{\textsf{http://cimg.sourceforge.net/} and
\textsf{http://www.kyb.tuebingen.mpg.de/bs/people/spider/}}.

\par We adopted a coarser representation of 9 bits per color for the $98,304$ pixels of each image, rather than the 24 available
ones to reduce the size of the RGB color space to $8^3=512$ from
the original set of $256^3=16,777,216$ colors. In this image retrieval experiment, we
used localization as the conditioning index set, dividing the
images into $1, 4, 4^2=16, 9$ and $9^2=81$ local
histograms (in Figure~\ref{fig:multiresgrid} the image was for instance
divided into $4^3=64$ windows). To define the branching process prior,
we simply set an uniform value over all the grid of $\varepsilon$ of $1/\alpha$, an usage
motivated by previous experiments led in a similar context~\cite{cuturi05context}. Finally, we
used kernels described in both~\cite{chapelle99svms} and~\cite
{hein05hilbertian} to define the base kernel $k$. These kernels can be
directly applied on sub-probability measures, which is not the case
for all kernels on multinomials, notably the Information Diffusion Kernel~\cite{Lafferty2005}. We report results for two families of kernels, namely the Radial Basis Function expressed for multinomials and
the entropy kernel based on the Jensen divergence~\cite{hein05hilbertian,cuturi05semigroup}:
$$
k_{a,b,\rho}(\theta,\,\theta')=e^{-\rho\sum|\theta_i^a-{\theta'_i}^a|^b}\;\; , \;k_h(\theta,\,\theta')=e^{-h\left(\frac{\theta+\theta'}{2}\right)+\frac{1}{2}\left(h(\theta)+h(\theta')\right)}.
$$
For most kernels not presented here, the multiresolution approach usually improved the performance in a similar way than the results presented in Table~\ref{tab:results}. Finally, we also report that using only the finest resolution available in each $(\alpha,D)$ setting, that is a branching process prior uniformly set to $1$, yielded better results than the use of the coarsest histogram without achieving however the same performance of the multiresolution averaging
framework, which highlights the interest of taking both coarse and fine perspectives into account. When $a=.25$ for instance, this setting produced 16.5\% and 16.2\% error rates for $\alpha=4$ and $D=1,2$, and 15.8\% for $\alpha=9$ and $D=1$.
\begin{table}[htbp]
\begin{center}
\begin{tabular}{|c||c|c|c|c|}
\hline
\multirow{2}{*}{Kernel} & \multicolumn{3}{|c|}{RBF, $b=1$, $\rho=.01$} & \multirow{2}{*}{JD}\\
& $a=.25$ & $a=.5$ & $a=1$ &\\
\hline
global histogram & 18.5 & 18.3 & 18.3 & 21.4 \\
$D=1,\alpha=4$ & 15.4 & 16.4 & 18.8 & 17 \\
$D=2,\alpha=4$ & \emph{13.9} & \emph{13.5} & 15.8 & 15.2 \\
$D=1,\alpha=9$ & 14.7 & 14.7 & 16.6 & 15 \\
$D=2,\alpha=9$ & 15.1 & 15.1 & 30.5 & 15.35\\
\hline\end{tabular}
\caption{Results for the Corel image database experiment in terms of error rate, with 4 fold cross-validation and 2 different types of tested kernels, the RBF and the Jensen Divergence.}\label{tab:results}
\end{center}
\end{table}
\section*{Acknowledgments}
MC would like to thank Jean-Philippe Vert and Arnaud Doucet for
fruitful discussions, as well as Xavier Dupr\'e for his help with
the CImg toolbox.
\bibliographystyle{plain}
\small\bibliography{bibliography}

\begin{thebibliography}{10}

\bibitem{stflour}
Olivier Catoni.
\newblock {\em Statistical learning theory and stochastic optimization, Ecole
  d'{\'e}t{\'e} de probabilit{\'e}s de Saint-Flour XXXI -2001}.
\newblock Number 1851 in Lecture Notes in Mathematics. Springer Verlag, 2004.

\bibitem{chapelle99svms}
O.~Chapelle, P.~Haffner, and V.~Vapnik.
\newblock Svms for histogram based image classification.
\newblock {\em IEEE Transactions on Neural Networks}, 10(5):1055, September
  1999.

\bibitem{cuturi05semigroup}
Marco Cuturi, Kenji Fukumizu, and Jean-Philippe Vert.
\newblock Semigroup kernels on measures.
\newblock {\em Journal of Machine Learning Research}, 6:1169--1198, 2005.

\bibitem{cuturi05context}
Marco Cuturi and Jean-Philippe Vert.
\newblock The context-tree kernel for strings.
\newblock {\em Neural Networks}, 18(8), 2005.

\bibitem{haussler99convolution}
David Haussler.
\newblock Convolution kernels on discrete structures.
\newblock Technical report, UC Santa Cruz, 1999.
\newblock USCS-CRL-99-10.

\bibitem{hein05hilbertian}
M.~Hein and O.~Bousquet.
\newblock Hilbertian metrics and positive definite kernels on probability
  measures.
\newblock January 2005.

\bibitem{jebara04probability}
Tony Jebara, Risi Kondor, and Andrew Howard.
\newblock Probability product kernels.
\newblock {\em Journal of Machine Learning Research}, 5:819--844, 2004.

\bibitem{joachims:2002a}
Thorsten Joachims.
\newblock {\em Learning to Classify Text Using Support Vector Machines:
  Methods, Theory, and Algorithms}.
\newblock Kluwer Academic Publishers, Dordrecht, 2002.

\bibitem{Lafferty2005}
John Lafferty and Guy Lebanon.
\newblock Diffusion kernels on statistical manifolds.
\newblock {\em Journal of Machine Learning Research}, 6:129--163, January 2005.

\bibitem{RaeSon04}
G.~R\"atsch and S.~Sonnenburg.
\newblock Accurate splice site prediction for caenorhabditis elegans.
\newblock In Bernhard Sch{\"o}lkopf, Koji Tsuda, and Jean-Philippe Vert,
  editors, {\em Kernel Methods in Computational Biology}. MIT Press, 2004.

\bibitem{schoelkopf02learning}
Bernhard Sch{\"o}lkopf and Alexander~J. Smola.
\newblock {\em Learning with Kernels: Support Vector Machines, Regularization ,
  Optimization, and Beyond}.
\newblock MIT Press, Cambridge, MA, 2002.

\bibitem{shimodaira02dynamic}
H.~Shimodaira, K.-I. Noma, M.~Nakai, and S.~Sagayama.
\newblock Dynamic time-alignment kernel in support vector machine.
\newblock In T.~G. Dietterich, S.~Becker, and Z.~Ghahramani, editors, {\em
  Advances in Neural Information Processing Systems 14}, Cambridge, MA, 2002.
  MIT Press.

\bibitem{tsuda02marginalized}
K.~Tsuda, T.~Kin, and K.~Asai.
\newblock Marginalized kernels for biological sequences.
\newblock {\em Bioinformatics}, 18(Suppl 1):268--275, 2002.

\bibitem{VerSaiAku04}
Jean-Philippe Vert, Hiroto Saigo, and Tatsuya Akutsu.
\newblock Local alignment kernels for protein sequences.
\newblock In Bernhard Sch{\"o}lkopf, Koji Tsuda, and Jean-Philippe Vert,
  editors, {\em Kernel Methods in Computational Biology}. MIT Press, 2004.

\end{thebibliography}
\end{document}